\documentclass[10pt,twocolumn,letterpaper]{article}

\usepackage{iccv}
\usepackage{times}
\usepackage{epsfig}
\usepackage{graphicx}
\usepackage{amsmath}
\usepackage{amssymb}
\usepackage{multirow}
\usepackage{soul}   
\usepackage[inline, shortlabels]{enumitem}
\usepackage[title]{appendix}

\newcommand{\eat}[1]{}

\usepackage[pagebackref=true,breaklinks=true,letterpaper=true,colorlinks,bookmarks=false]{hyperref}

\iccvfinalcopy 

\def\iccvPaperID{4151} 

\ificcvfinal\fi
\begin{document}

\title{A Fast and Accurate One-Stage Approach to Visual Grounding}

\author{Zhengyuan Yang$^2$\thanks{Work done while Z.Yang was an intern at Tencent AI Lab at Bellevue.
} \quad Boqing Gong$^1$\thanks{Now at Google.} \quad Liwei Wang$^1$ \quad Wenbing Huang$^1$ \quad Dong Yu$^1$ \quad Jiebo Luo$^2$\\
$^1$Tencent AI Lab \qquad\qquad $^2$University of Rochester\\
{\tt\small \{zyang39, jluo\}@cs.rochester.edu, boqinggo@outlook.com}\\
{\tt\small \{liweiwang, dongyu\}@tencent.com, hwenbing@126.com}}

\maketitle

\begin{abstract}
   We propose a simple, fast, and accurate one-stage approach to visual grounding, inspired by the following insight. The performances of existing  propose-and-rank two-stage methods are capped by the quality of the region candidates they propose in the first stage --- if none of the candidates could cover the ground truth region, there is no hope in the second stage to rank the right region to the top. To avoid this caveat, we propose a one-stage model that enables end-to-end joint optimization. The main idea is as straightforward as fusing a text query's embedding into the YOLOv3 object detector, augmented by spatial features so as to account for spatial mentions in the query. Despite being simple, this one-stage approach shows great potential in terms of both accuracy and speed for both phrase localization and referring expression comprehension, according to our experiments. Given these results along with careful investigations into some popular region proposals, we advocate  for  visual grounding  a paradigm shift from the conventional two-stage methods to the one-stage framework. 
\end{abstract}


\vspace{-0.1in}
\section{Introduction}
We propose a simple, fast, and accurate one-stage approach to visual grounding, which aims to ground a natural language query (phrase or sentence) about an image onto a correct region of the image. By defining visual grounding at this level, we deliberately abstract away the subtle distinctions between phrase localization~\cite{plummer2017flickr30k,wang2019learning}, referring expression comprehension~\cite{kazemzadeh2014referitgame,mao2016generation,yu2016modeling,yu2018mattnet,liu2019improving}, natural language object retrieval~\cite{hu2016natural,li2017deep}, visual question segmentation~\cite{gan2017vqs,hu2016segmentation,liu2017recurrent,margffoy2018dynamic}, etc., each of which can be seen as a variation of the general visual grounding problem. We benchmark our one-stage approach for both phrase localization and referring expression comprehension. Results show that it is about 10 times faster than the state-of-the-art two-stage methods and meanwhile more accurate than them. Hence, we expect this work provides for visual grounding a new strong baseline, upon which one can conveniently build further to tackle variations (e.g., phrase localization) to the basic visual grounding problem by bringing in corresponding domain knowledge (e.g., attributes, relationship between phrases, spatial configuration of regions, etc.). 

\begin{figure}[t]
\begin{center}
   \centerline{\includegraphics[width=8cm]{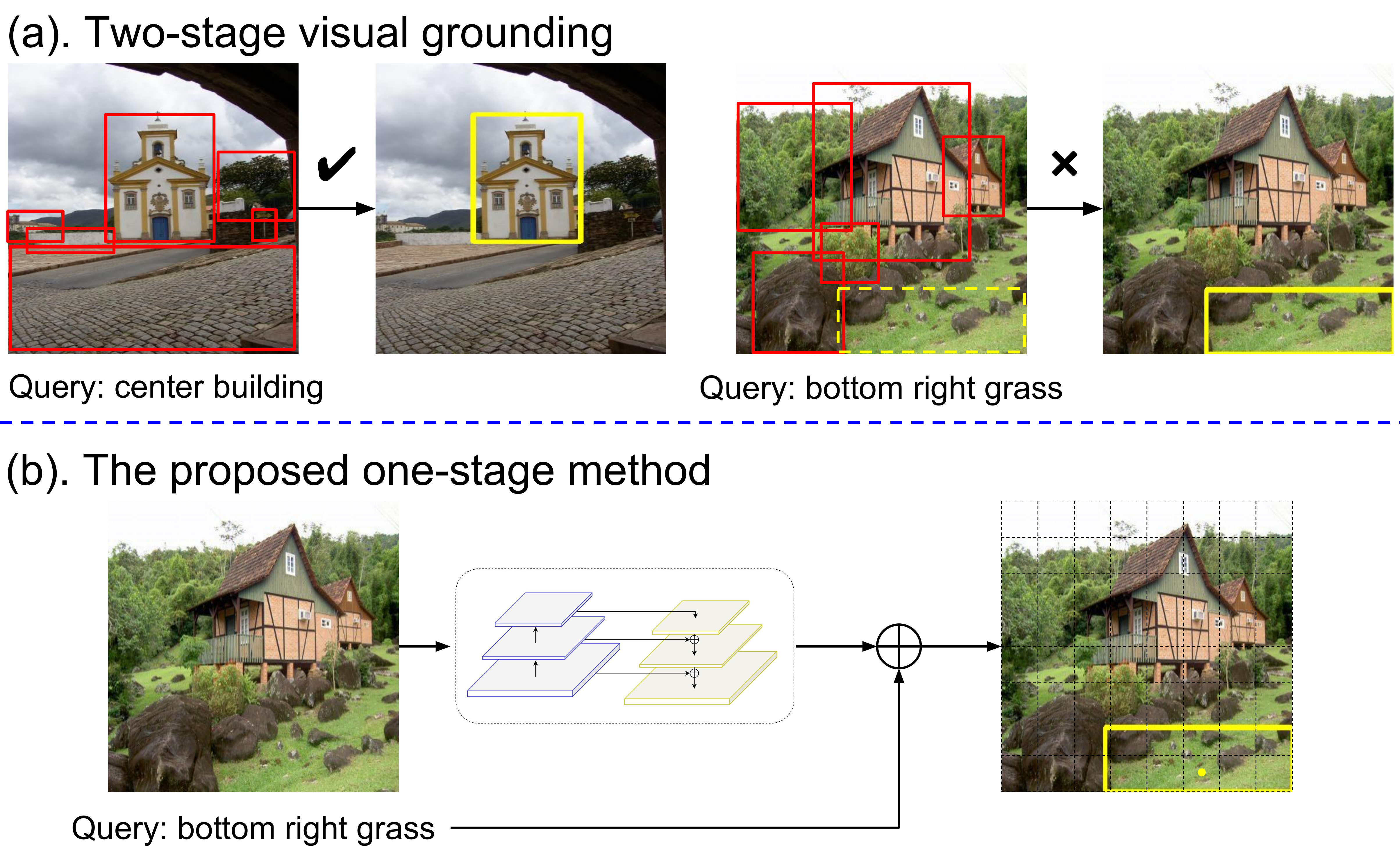}}
\end{center}
\vspace{-0.3in}
	\caption{Visual grounding is the task of localizing a language query in an image. The output is often a bounding box as drawn in the yellow color. {\bf (a).} Existing two-stage methods first extract region candidates and then rank them according to their similarities with the query. The inference speed is slow and the performance is capped by the quality of the region proposals (e.g., on the right, the ``bottom right grass'' is not covered by any of the region candidates). {\bf (b).} Our proposed one-stage method directly predicts a grounding box given the input image and a query. It is hence significantly faster and also accurate in inference.}
\vspace{-0.2in}
\label{fig:intro}
\end{figure}

Visual grounding is key to machine intelligence and provides a natural channel for humans to communicate with machines about the physical world. Its potential applications include but are not limited to robotics, human-computer interaction, and early education. In addition, a good visual grounding model can benefit a variety of research problems such as visual question answering~\cite{zhu2016visual7w,gan2017vqs,li2018tell}, image captioning~\cite{you2016image,anderson2018bottom,feng2018unsupervised}, and image retrieval~\cite{salvador2016faster}. 

There are mainly two thriving threads of work in visual grounding: phrase localization~\cite{kazemzadeh2014referitgame,plummer2017flickr30k,wang2019learning} and referring expression comprehension~\cite{mao2016generation,yu2016modeling,yu2018mattnet,hu2016natural,li2017deep} --- plus some work on grounding as segmentation~\cite{hu2016segmentation,gan2017vqs,liu2017recurrent,margffoy2018dynamic}. The language query in the former is a local phrase of a full sentence describing an image, implying that multiple phrase queries could co-occur in the sentence. In the latter, the query is an expression referring to a particular region of an image through a combination of object categories, attributes, relationships with other objects, etc. Notably, in phrase localization, an image region linked to a phrase of one sentence can also be linked to a phrase of another sentence, establishing a coreference chain. Compared with phrase localization, referring expression has less ambiguity in general. 

Recent advances in computer vision and natural language processing offer a rich set of tools, such as region proposals~\cite{zitnick2014edge,uijlings2013selective}, object detection~\cite{girshick2015fast,ren2015faster,he2017mask}, text embedding~\cite{perronnin2010improving,mikolov2013distributed,devlin2018bert}, syntactic parsing~\cite{socher2013parsing}, etc., leading to methods~\cite{wang2016learning,wang2019learning,plummerCITE2018,chen2017msrc,yu2017joint,yu2018mattnet} exploiting various cues in the visual grounding problem. However, somehow surprisingly, the main bodies of these methods are remarkably alike: they propose multiple region candidates per image and then rank them according to their similarities with the language query. We contend that this propose-and-rank two-stage framework is flawed in at least two major ways. 
\vspace{-6pt}
\begin{itemize} \setlength\itemsep{-3pt}
\item If none of the region candidates of the first stage hits the ground truth region, the whole framework fails no matter how good the second ranking stage could perform. We find that 200 Edgebox region proposals~\cite{zitnick2014edge} per image can only hit 68\% of the ground truth regions in ReferItGame~\cite{kazemzadeh2014referitgame}, a benchmark dataset for referring expression comprehension. A hit is considered successful if any of the 200 proposals could reach 0.5 or higher intersection-over-union (IoU)~\cite{plummer2017flickr30k} with the ground truth region. 
\item Most of the computation spent on the region candidates, such as generating proposals, extracting features, fusing with the query embedding, scoring similarities, and so on, are merely to rank them down to the list. After all, in most test cases, only one or two region proposals are correct. We believe this scheme is a waste of computation and should be improved.
\vspace{-6pt}
\end{itemize} 

The two caveats are left unresolved probably due to the long-standing pursuit of how to model different cues in visual grounding. In this paper, we take a step back and re-examine the visual grounding problem at an abstract level, without discriminating the query types. We propose to shift the paradigm from grounding as ranking multiple region candidates to directly proposing one region as the output. 

To this end, we study an end-to-end one-stage approach to visual grounding. The main idea is as straightforward as fusing a text query's embedding into the YOLOv3 object detector~\cite{redmon2018yolov3}. Additionally, we augment the feature maps with spatial features to account for spatial mentions in the language queries (e.g., ``the man on the right''). Finally, we replace the sigmoid output layer with a softmax function in order to enforce the network to generate only one image region in response to a query. Other cues explored in the two-stage methods, such as attributes, attention, bounding box annotations around extra objects, and so on, can be naturally added to our  one-stage model. We focus on the vanilla model in the main text and examine its extensibilities to some other cues in supplementary materials.

The advantages of this one-stage approach are multiple-fold. First of all, it is fast in inference. It extracts features from the input image with only one pass and then directly predicts the coordinates of the output region. Without any code optimization, our implementation is about 10 times faster than state-of-the-art two-stage methods. Additionally, it is also accurate. Unlike the two-stage framework whose performance is capped by the region candidates, it enables end-to-end optimization. We show promising results on both phrase localization and referring expression comprehension. Finally, it generalizes better to different datasets than the two-stage methods because it does not depend on any additional tools or pre-trained models. Hence, we advocate this one-stage framework for future work on visual grounding and hope our approach in this work provides a new strong baseline.




\section{Approach}
In this section, we first review the existing two-stage frameworks for visual grounding~\cite{wang2019learning,plummer2017flickr30k,yu2016modeling,mao2016generation,yu2018mattnet,rohrbach2016grounding,plummerCITE2018} and then present our one-stage approach in detail. 

\subsection{Two-stage methods}
Conventional methods for visual grounding, especially for the task of phrase localization~\cite{plummer2017flickr30k,wang2019learning,plummerCITE2018,chen2017msrc}, are mainly composed of two separate stages. As shown in Figure~\ref{fig:intro}, given an input image, the first step is to generate candidate regions using either unsupervised object proposal methods~\cite{zitnick2014edge,wang2019learning,plummerCITE2018,chen2017msrc} or a pre-trained object detection network~\cite{zhang2018grounding,yu2018mattnet}. The second step is to rank the candidate regions conditioning on a language query about the image. Most existing two-stage methods differ from each other in the second step by scoring functions, network architectures, multi-task learning, and training algorithms. 
A number of studies~\cite{zhang2017discriminative, wang2019learning} cast the second step as a binary classification task, where a region-query pair is tagged ``positive'', ``negative'', or ``ignored'' based on the region's IoU with the ground truth region. The maximum-margin ranking loss is another popular choice for the second stage~\cite{mao2016generation, nagaraja2016modeling, wang2016learning}. 

As a concrete example, we next describe the similarity network~\cite{wang2019learning,plummerCITE2018} since it gives rise to state-of-the-art performance on benchmark datasets. The authors employ a Fast R-CNN~\cite{girshick2015fast,ren2015faster} pre-trained on Pascal~\cite{everingham2010pascal} to extract visual features for each candidate region. To embed the text query, they find the Fisher encoding~\cite{perronnin2010improving} works as well as or better than recurrent neural networks. The region features and query embeddings are fed through two network branches, respectively, before they merge by a layer of element-wise multiplication. A few nonlinear layers are added after they merge. Finally, the network outputs a similarity score by a sigmoid function. The authors train this network by a cross-entropy loss with positive labels for the matched pairs of regions and queries and negative labels for the mismatched pairs. A region is a match to the query if its IoU with the ground truth is greater than 0.7 and the regions with IoUs less than 0.3 are considered mismatches. 

The overall performance of the two-stage framework is capped by the first stage. Besides, the candidate regions cause heavy computation cost. We next present a different paradigm, a one-stage visual grounding network which enables end-to-end optimization and is both fast and accurate.

\subsection{Our one-stage approach}
\label{sec:method}
In short, our one-stage approach to the visual grounding is to fuse a text query's embedding into YOLOv3~\cite{redmon2018yolov3}, augment it with spatial features as the spatial configuration is frequently used by a query, replace its sigmoid output layer with a softmax function because we only need to return one region for a query, and finally train the network with YOLO's loss~\cite{redmon2016you}. Despite being simple, this one-stage method signifies a paradigm shift away from the prevalent two-stage framework, and it gives rise to superior results in terms of both accuracy and speed. 

\begin{figure*}[t]
\begin{center}
   \centerline{\includegraphics[width=16cm]{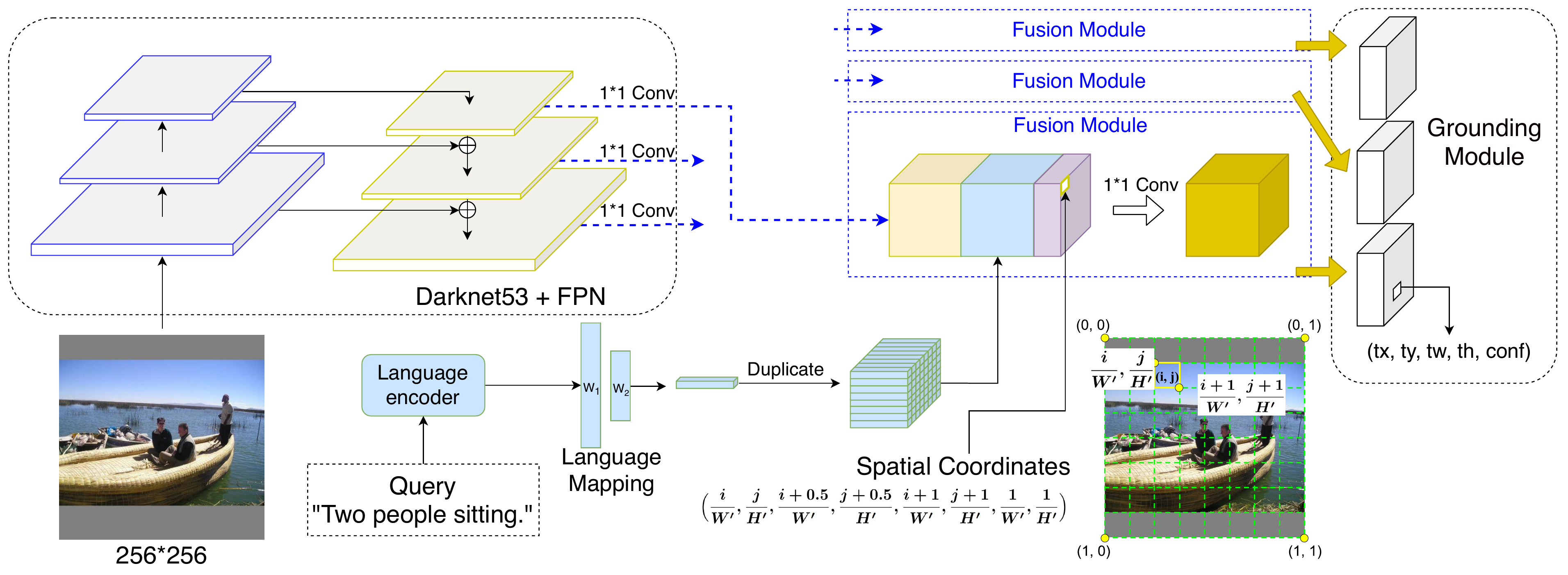}}
\end{center}
\vspace{-0.3in}
	\caption{The proposed end-to-end one-stage visual grounding framework.}
\label{fig:arch}
\vspace{-0.15in}
\end{figure*}


We present this vanilla one-stage model as below and amend it in supplementary materials to account for some  cues explored in the two-stage methods. Figure~\ref{fig:arch} illustrates the network architecture, mainly consisting of three feature encoding modules and three fusion modules. 

\vspace{3pt}
\noindent \textbf{Visual and text feature encoding.} Our model is end-to-end, taking as input an image and a text query and then returning an image region as the response to the query. For the text query, we embed it to a 768D real-valued vector using the uncased version of Bert~\cite{devlin2018bert}, followed by two fully connected layers either with 512 neurons. In addition, we also test the other embedding methods to fairly compare with the existing works. In particular, recent works~\cite{plummer2017flickr30k,wang2019learning,plummerCITE2018} employ Fisher vectors of word2vec~\cite{perronnin2010improving,mikolov2013distributed}. Bidirectional LSTMs are adopted in~\cite{chen2017msrc,rohrbach2016grounding}.

We use Darknet-53~\cite{redmon2018yolov3} with feature pyramid networks~\cite{lin2017feature} to extract visual features for the input image, which is resized to $256\times 256$, at three spatial resolutions: $8\times 8\times D_1$, $16\times 16\times D_2$, and $32\times 32\times D_3$. In other words, the feature maps are respectively $\frac{1}{32}$, $\frac{1}{16}$, and $\frac{1}{8}$ of the original image size. There are $D_1=1024, D_2=512, \text{and } D_3=256$ feature channels at the three resolutions, respectively. We add a $1\times 1$ convolution layer with batch normalization and RELU to map them all to the same dimension $D=512$.

\vspace{3pt}
\noindent \textbf{Spatial feature encoding.} We find that the text queries often use spatial configurations to refer to objects, such as ``the man on the left'' and ``the bottom right grass''. However, the Darknet-53 features mainly capture visual appearances, lacking the position information. Hence, we explicitly encode some spatial features for each position of the three spatial resolutions.  Specifically, as shown in Figure~\ref{fig:arch}, we generate a coordinate map of size $W'\times H'\times D_\text{spatial}$ at each  resolution, where $W'$ and $H'$ are the spatial size of a visual feature map, i.e., $8\times 8$, $16\times 16$, or $32\times 32$, and $D_\text{spatial}=8$ indicating we encode eight spatial features. If we place the feature map in a coordinate system such that its top-left and bottom-right corners lie at $(0,0)$ and $(1,1)$, respectively, the eight features for any position $(i,j)$, $i \in \{0,1,\cdots, W'-1\}$ and $j\in \{0,1,\cdots, H'-1\}$, are calculated as follows: 
\begin{align}
\Big(\frac{i}{W'}, \frac{j}{H'}, \frac{i+0.5}{W'}, \frac{j+0.5}{H'}, \frac{i+1}{W'}, \frac{j+1}{H'}, \frac{1}{W'}, \frac{1}{H'}\Big), \notag
\end{align}
which captures the coordinates of the top-left corner, center, and bottom-right corner of the grid at $(i,j)$, along with the inverse of $W'$ and $H'$. 

\vspace{3pt}
\noindent \textbf{Fusion.} We use the same operation to fuse the visual, text, and spatial features at the three spatial resolutions. In particular, we first broadcast the  query embedding to each spatial location $(i,j)$ and then concatenate it with the visual and spatial features, giving rise to a $512+512+8=1032\text{D}$ feature vector. The visual, text, and spatial features are $\ell_2$ normalized respectively before the concatenation. We add a $1\times 1$ convolution layer to better fuse them at each location independently. We have also tested $3\times 3$ convolution kernels hoping to make the fusion aware of the neighborhood structure, and yet the results are about the same as the $1\times 1$ fusion. After this fusion step, we have a 512D feature vector for each location of the three spatial resolutions, i.e., three feature blobs of the sizes $8\times 8 \times 512$, $16\times 16 \times 512$, and $32\times 32\times 512$, respectively.


\vspace{3pt}
\noindent \textbf{Grounding.}
The grounding module takes the fused features as input and generates a box prediction to ground the language query onto an image region. We design this module by following YOLOv3's output layer except that we 1) re-calibrate its anchor boxes and 2) change its sigmoid layer to a softmax function. 

There are $8\times 8+16\times 16+32\times 32=1344$ locations out of the three spatial resolutions --- and each location is associated with a 512D feature vector as a result of the fusion module. YOLOv3 centers around  each of the locations three anchor boxes. To better fit our grounding datasets, we customize the widths and heights of the anchors by $K$-means clustering over all the ground truth grounding boxes in the training set with $(1-\text{IoU})$ as the distance~\cite{redmon2016yolo9000,redmon2018yolov3}. 

There are $(3\text{ anchors per location}\times 1344 \text{ locations})=4032$ anchor boxes in total. What YOLOv3 predicts is, out of each anchor box, four quantities by regression for shifting the center, width, and height of the anchor box, along with the fifth quantity by a sigmoid function about the confidence on this shifted box. We keep the regression branch as is. As only one region is desired as the output for grounding the  query --- at least according to the current formalization of the visual grounding problem, we replace the sigmoid functions with a softmax function over all the 4032 boxes. Accordingly, we replace the loss function over the confidence scores by a cross entropy between this softmax  and a one-hot vector --- the anchor box which has the highest IoU with the ground truth region is labeled $1$ and all the others are labeled  $0$. We refer readers to~\cite{redmon2018yolov3} for more details.

\subsection{Comparison to other one-stage grounding work}
We contrast our approach to some closely related works, including two existing one-stage grounding methods~\cite{zhao2018weakly,yeh2017interpretable} and some on grounding as segmentation~\cite{hu2016segmentation,liu2017recurrent,margffoy2018dynamic,gan2017vqs}.



The Interpretable and Globally Optimal Prediction (IGOP)~\cite{yeh2017interpretable} also tries to solve supervised visual grounding in a one-stage manner. IGOP employs feature maps from multiple vision tasks (e.g., object detection, semantic segmentation, pose estimation, etc.) and models the phrase localization task as finding a box on the feature maps which encapsulates the smallest energy. Since IGOP relies on multiple extra pre-trained vision models, it is not clear how to optimize it  end-to-end.

The Multiple-scale Anchored Transformer Network (MATN)~\cite{zhao2018weakly} is also a one-stage grounding model. However, many design consideratons of this network are to account for weakly supervised visual grounding. Besides, MATN directly predicts a single box as the output, essentially searches for one box out of a huge search space at the scale $O(W^2H^2)$, where $W, H$ are width and height of the input image. This scheme has been shown inferior to those based on anchor boxes in object detection~\cite{redmon2016yolo9000,redmon2018yolov3}, unless one has sufficiently big training sets.


We also briefly discuss some works on grounding text queries to segmentation masks~\cite{hu2016segmentation,liu2017recurrent,margffoy2018dynamic,gan2017vqs}. Due to the irregular shapes of the segmentation masks, it is hard to follow the propose-and-rank two-stage framework to output segmentation masks. Instead, they naturally employ one-stage frameworks. However, their network architectures, especially the output layer, are very different from ours. 

\section{Experiments}



\vspace{-2pt}
\subsection{Datasets and experiment protocols}
\vspace{-2pt}
We evaluate the proposed one-stage visual grounding approach on the Flickr30K Entities dataset~\cite{plummer2017flickr30k} and the ReferItGame dataset~\cite{kazemzadeh2014referitgame}. The supplementary materials contain additional results on RefCOCO~\cite{yu2016modeling}. Flickr30K Entities augments the original Flickr30K~\cite{young2014image} with region-phrase correspondence annotations. It links 31,783 images in Flickr30K with 427K referred entities. We follow the same training/validation/test split used in the previous work~\cite{plummer2017flickr30k} in our experiments. ReferItGame~\cite{kazemzadeh2014referitgame} has 20,000 images from the SAIAPR-12 dataset~\cite{escalante2010segmented}. We employ a cleaned version of the split provided by~\cite{hu2016natural}, which has 9,000, 1,000 and 10,000 images in the training, validation, and test sets, respectively. Following the same evaluation protocol in prior works~\cite{plummer2017flickr30k, rohrbach2016grounding},  given a language query, an output image region is considered correct if its IoU is at least 0.5 with the ground truth bounding box. 

\vspace{3pt}
\noindent \textbf{Some details of our model architecture.}
We use Darknet-53~\cite{redmon2018yolov3} pre-trained on COCO object detection~\cite{lin2014microsoft} as the visual encoder. To embed the language queries, we test Bert~\cite{devlin2018bert}, a bi-LSTM framework used in~\cite{chen2017msrc}, and a Fisher vector encoding used in~\cite{plummer2017flickr30k,wang2019learning,plummerCITE2018}. 
We generate the anchor boxes by $K$-means clustering following the procedure of~\cite{redmon2016yolo9000,redmon2018yolov3}. The anchors on ReferitGame are $(18\times22), (48\times28), (29\times52), (91\times48), (50\times91), (203\times57), (96\times127), (234\times100), (202\times175)$ and on Flickr30K Entities are $(17\times16), (33\times35), (84\times43), (50\times74), (76\times126), (125\times81), (128\times161), (227\times104), (216\times180)$.

\vspace{3pt}
\noindent \textbf{Training details.}
We keep the original image ratio when we resize an input image. We resize its long edge  to 256 and then pad the image pixels' mean value along the short edge so that the final image size is $256\times256$. We follow~\cite{redmon2018yolov3} for data augmentation, i.e., adding randomization to the color space (saturation and intensity), horizontal flip, and random affine transformations. We train the model with RMSProp~\cite{tieleman2012lecture} optimization. We start with a learning rate of $10^{-4}$ and follow a polynomial schedule with a power of 1. As the Darknet has been pre-trained, we multiply the main learning rate by 0.1 for the Darknet portion of our model. The batch size is 32 in all our experiments. We observe about 1\% improvement when we use larger batch sizes on a workstation with eight P100 GPUs, but we opt to report the results of the small batch size (32) so that one can easily reproduce our results on a desktop with two GPUs.


\vspace{3pt}
\noindent \textbf{Competing baselines beyond existing methods.}
We compare with state-of-the-art visual grounding methods, about which the descriptions are deferred to Section~\ref{sec:result}. Beyond them, we also systematically study the following baselines and variations of our approach.
\vspace{-6pt}

\begin{itemize}
\setlength\itemsep{-3pt}
\item{\bf Similarity Net-Darknet.}
Previous two-stage methods often use detection networks with a VGG-16 backbone~\cite{simonyan2014very} to extract visual features, while Darknet is adopted in our model. Naturally, one may wonder about the influence of the backbones in addition to the framework change from the two stages to the one stage. To single out the influence of the backbone networks, we construct a baseline using the two-stage similarity network~\cite{wang2019learning} based on the Darknet visual features, modifying the code released with~\cite{plummerCITE2018}.  We first pool region features from all three feature blobs output by the feature pyramid network in YOLOv3, then $\ell_2$ normalize them, respectively, and finally concatenate them as the visual features. 

\item{\bf Similarity Net-Resnet.}
We also test in the similarity network visual features extracted by Mask R-CNN~\cite{he2017mask} with a Resnet-101~\cite{he2016deep} backbone, which is pre-trained on COCO detection. The feature dimension is 2048. 

\item{\bf CITE-Resnet.}
Furthermore, we compare to CITE~\cite{plummerCITE2018} with Resnet-101 features. We keep the number of embeddings as the default value $K=4$ in CITE. Region proposals and visual and language encoders remain the same as ``Similarity Net-Resnet''.

\item{\bf Ours-FV.}
The Fisher vector (FV) encoding~\cite{perronnin2010improving} of word2vec~\cite{mikolov2013distributed} features is used in some state-of-the-art visual grounding methods~\cite{plummer2017flickr30k,wang2019learning,plummerCITE2018}. We include it in our approach as well. A language query is encoded to a 6000D FV embedding. 

\item{\bf Ours-LSTM.}
The LSTM encoding of language queries is also frequently used in the literature~\cite{chen2017msrc,rohrbach2016grounding}, so we investigate its effect in our approach as well. We use a bi-LSTM layer with 512D hidden states in this work. We do not use word2vec features to initialize the embedding layer.

\item{\bf Ours-Bert.}
We use the uncased version of Bert~\cite{devlin2018bert} that outputs a 768D embedding as our main language query encoder. We do not update the Bert parameters  during training.

\item{\bf Ours-Bert-no Spatial.}
In this ablated version of our approach, we remove the spatial features and only fuse the visual and text features. 

\vspace{-6pt}
\end{itemize}

\begin{table*}[t]
\centering
\caption{Phrase localization results on the test set of Flickr30K Entities~\cite{plummer2017flickr30k}.}
\vspace{-0.0in}
\begin{tabular}{ l l l l c c }
    \hline
    \small{Method} & \small{Region Proposals} & \small{Visual Features} & \small{Language Embedding} & \small{Accu@0.5} & \small{Time (ms)}\\
    \hline
    SCRC~\cite{hu2016natural} & Edgebox N=100 & VGG16-Imagenet & LSTM & 27.80 & - \\
    DSPE~\cite{wang2016learning} & Edgebox N=100 & VGG19-Pascal & Word2vec, FV & 43.89 & - \\
    GroundeR~\cite{rohrbach2016grounding} & Selec. Search N=100 & VGG16-Pascal & LSTM & 47.81 & - \\
    CCA~\cite{plummer2017flickr30k} & Edgebox N=200 & VGG19-Pascal & Word2vec, FV & 50.89 & - \\
    IGOP~\cite{yeh2017interpretable} & None & Multiple Network & N-hot & 53.97 & - \\
    MCB + Reg + Spatial~\cite{chen2017msrc} & Selec. Search N=100 & VGG16-Pascal & LSTM & 51.01 & - \\
    MNN + Reg + Spatial~\cite{chen2017msrc} & Selec. Search N=100 & VGG16-Pascal & LSTM & 55.99 & - \\
    Similarity Net~\cite{wang2019learning} & Edgebox N=200 & VGG19-Pascal & Word2vec, FV & 51.05 & -  \\
    Similarity Net by CITE~\cite{plummerCITE2018} & Edgebox N=200 & VGG16-Pascal & Word2vec, FV & 54.52 & -  \\
    CITE~\cite{plummerCITE2018} & Edgebox N=500 & VGG16-Pascal & Word2vec, FV & 59.27 & - \\
    CITE~\cite{plummerCITE2018} & Edgebox N=500 & VGG16-Flickr30K & Word2vec, FV & 61.89 & - \\ 
    \hline
    Similarity Net-Resnet~\cite{wang2019learning} & Edgebox N=200 & Res101-COCO & Word2vec, FV & 60.89 & 184\\
    CITE-Resnet~\cite{plummerCITE2018} & Edgebox N=200 & Res101-COCO & Word2vec, FV & 61.33 & 196\\
    Similarity Net-Darknet~\cite{wang2019learning} & Edgebox N=200 & Darknet53-COCO & Word2vec, FV & 41.04 & 305\\
    \hline
    Ours-FV & None & Darknet53-COCO & Word2vec, FV & 68.38 & {\bf 16} \\
    Ours-LSTM & None & Darknet53-COCO & LSTM & 67.62 & 21 \\
    Ours-Bert-no Spatial & None & Darknet53-COCO & Bert & 67.08 & 38  \\
    Ours-Bert & None & Darknet53-COCO & Bert & {\bf 68.69} & 38 \\
    \hline
\end{tabular}
\vspace{-0.0in}
\label{table:flickr}
\end{table*}

\vspace{-2pt}
\subsection{Visual grounding results}
\vspace{-2pt}
\label{sec:result}
\noindent {\bf Flickr30K Entities.} 
Table~\ref{table:flickr} reports the phrase localization results on the Flickr30K Entities dataset. The \textbf{top portion} of the table contains the numbers of several state-of-the-art visual grounding methods~\cite{hu2016natural,wang2016learning,rohrbach2016grounding,plummer2017flickr30k,yeh2017interpretable,chen2017msrc,wang2019learning,plummerCITE2018}. The results of two additional versions of the similarity network~\cite{wang2019learning}, respectively based on Resnet and Darkent, are shown in the \textbf{middle} of the table. Finally, the four rows at the \textbf{bottom} are different variations of our own approach. 

We list in the ``Region Proposals'' column different region proposal techniques used in the visual grounding methods, followed by the number of region candidates per image (e.g., N=100). Edgebox~\cite{zitnick2014edge} and selective search~\cite{uijlings2013selective} are two popular options for proposing the regions. In the ``Visual Features'' column, we list the backbone networks followed by the datasets on which they are pre-trained. The ``Language Embedding'' column indicates the query embedding adopted by each grounding method.

Among the two-stage methods, not surprisingly,  ``Similarity Net-Resnet'' gives much better results than ``Similarity Net'' because the Resnet visual features are generally higher-quality than the VGG features. 

Although Darknet-53 and Resnet-101 give rise to  comparable results on ImageNet~\cite{ILSVRC15}, the Darknet features lead to poor visual grounding results. This is reasonable because Darknet does not have a separate region proposal network, making it tricky to extract the region features. Furthermore, the large down-scale ratios (1/8, 1/16, and 1/32) and the low feature dimensions (256, 512, 1024) of Darknet make its region features not as discriminative as Resnet's. 

Our one-stage method and its variations outperform the two-stage approaches with large margins.  By the last two rows of the table, we investigate the effectiveness of our spatial features. It is clear that the spatial information boosts the accuracy of ``Ours-Bert-no Spatial'' by about $1.6\%$. Finally, we note that language embedding techniques only slightly influence the results within a small range. 




\begin{table*}[t]
\centering
\caption{Referring expression comprehension results on the test set of  ReferItGame~\cite{kazemzadeh2014referitgame}.}
\vspace{-0.0in}
\begin{tabular}{ l l l l c c }
    \hline
    \small{Method} & \small{Region Proposals} & \small{Visual Features} & \small{Language Embedding} & \small{Accu@0.5} & \small{Time (ms)}\\
    \hline
    SCRC~\cite{hu2016natural} & Edgebox N=100 & VGG16-Imagenet & LSTM & 17.93 & - \\
    GroundeR + Spacial~\cite{rohrbach2016grounding} & Edgebox N=100 & VGG16-Pascal & LSTM & 26.93 & - \\
    VC~\cite{zhang2018grounding} & SSD Detection~\cite{liu2016ssd} & VGG16-COCO & LSTM &  31.13 & - \\
    CGRE~\cite{luo2017comprehension} & Edgebox & VGG16 & LSTM &  31.85 & - \\
    MCB + Reg + Spatial~\cite{chen2017msrc} & Edgebox N=100 & VGG16-Pascal & LSTM & 26.54 & - \\
    MNN + Reg + Spatial~\cite{chen2017msrc} & Edgebox N=100 & VGG16-Pascal & LSTM & 32.21 & - \\
    Similarity Net by CITE~\cite{plummerCITE2018} & Edgebox N=500 & VGG16-Pascal & Word2vec, FV & 31.26 & - \\
    CITE~\cite{plummerCITE2018} & Edgebox N=500 & VGG16-Pascal & Word2vec, FV & 34.13 & - \\
    IGOP~\cite{yeh2017interpretable} & None & Multiple Network & N-hot & 34.70 & - \\
    \hline
    Similarity Net-Resnet~\cite{wang2019learning} & Edgebox N=200 & Res101-COCO & Word2vec, FV & 34.54 & 184\\
    CITE-Resnet~\cite{plummerCITE2018} & Edgebox N=200 & Res101-COCO & Word2vec, FV & 35.07 & 196\\
    Similarity Net-Darknet~\cite{wang2019learning} & Edgebox N=200 & Darknet53-COCO & Word2vec, FV & 22.37 & 305 \\
    \hline
    Ours-FV & None & Darknet53-COCO & Word2vec, FV & 59.18 & {\bf 16} \\
    Ours-LSTM & None & Darknet53-COCO & LSTM & 58.76 & 21 \\
    Ours-Bert-no Spatial & None & Darknet53-COCO & Bert & 58.16 & 38 \\
    Ours-Bert & None & Darknet53-COCO & Bert & {\bf 59.30} & 38 \\
    \hline
\end{tabular}
\vspace{-0.1in}
\label{table:clef}
\end{table*}

\vspace{3pt}
\noindent {\bf ReferitGame.} 
Table~\ref{table:clef} reports the referring expression comprehension results on ReferItGame~\cite{kazemzadeh2014referitgame}. Organizing the results in the same way as Table~\ref{table:flickr}, the top portion of the table is about state-of-the-art grounding methods~\cite{hu2016natural,luo2017comprehension,zhang2018grounding,rohrbach2016grounding,yeh2017interpretable,chen2017msrc,wang2019learning,plummerCITE2018}, the middle  is two versions of the similarity network, and the bottom shows our results. 

We draw from Table~\ref{table:clef} about the same observation as from Table~\ref{table:flickr}. In general, our model with Darknet visual features and Bert query embeddings outperforms the existing methods by large margins. Careful analyses reveal that the poor region candidates in the first stage are a major reason that the two-stage methods underperform ours. We present these analyses in Section~\ref{sec:oracle}.

\vspace{-3pt}
\subsection{Inference time comparison}
\vspace{-2pt}
A fast inference speed is one of the major advantages of our one-stage method. We list the inference time in the rightmost columns of Tables~\ref{table:flickr}~and~\ref{table:clef}, respectively. We conduct all the tests on a desktop with Intel Core i9-9900K@3.60GHz and NVIDIA 1080TI. Typical two-stage approaches generally take more than 180ms to process one image-query pair, and they spend most of the time on generating region candidates and extracting features for them. In contrast, our one-stage approaches all take less than 40ms to ground a language query to an image --- especially, ``Ours-FV'' takes only 16ms, making it potentially feasible for real-time applications. 

\vspace{-2pt}
\subsection{Oracle analyses about the region candidates}
\vspace{-2pt}
\label{sec:oracle}
Why could the one-stage methods achieve those big improvements over the two-stage ones? We conjecture that it is mainly because our one-stage framework can avoid imperfect region candidates. In contrast, the performances of the two-stage methods are capped by the \textbf{hit rate} of the region candidates they propose in the first stage. We say a ground truth region is hit by the region candidates if its IoU is greater than 0.5 with any of the candidates, and the hit rate is the number of ground truth regions hit by the candidates divided by the total number of ground truth regions. 


We study the hit rates of some popular region proposal methods: Edgebox~\cite{zitnick2014edge}, selective search~\cite{uijlings2013selective}, the region proposal network in Mask R-CNN~\cite{he2017mask} pre-trained on COCO~\cite{he2017mask}, Mask R-CNN itself whose detection results are regarded as region candidates, and our one-stage approach whose box predictions are considered the region candidates. We keep top N=200 region candidates for each of them or as many regions as possible if it outputs less than 200 regions.


Table~\ref{table:oracle} shows the hit rates on both Flickr30K Entities and ReferItGame. It is interesting to see that the hit rates are in general higher on Flickr30 than on ReferItGame, especially when Edgebox generated proposals are used, somehow explaining why the two-stage grounding results on Flickr30K Entities (Table~\ref{table:flickr}) are better than those on ReferItGame (Table~\ref{table:clef}). Another notable observation is that the top 200 boxes of our approach have much higher hit rates than the other techniques, verifying the benefit of learning in an end-to-end way. 

\begin{figure}[t]
\begin{center}
   \centerline{\includegraphics[width=7.5cm]{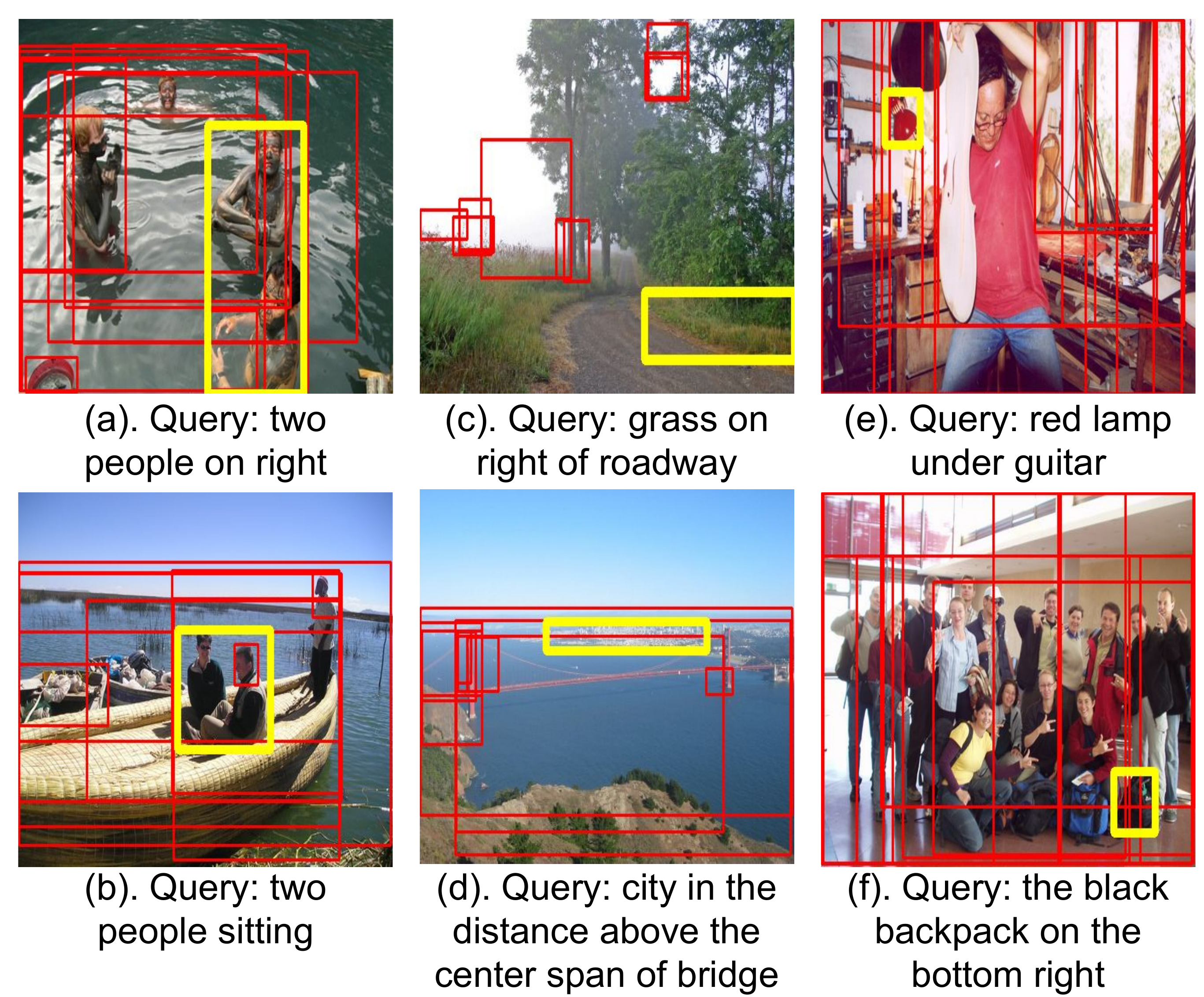}}
\end{center}
\vspace{-0.3in}
	\caption{Failure cases of the Edgebox region candidates (boxes shown in the red color) on ReferitGame. The yellow boxes are ground truths. For visualization purpose, we randomly hide some region candidates.}
\vspace{-0.2in}
\label{fig:edgebox}
\end{figure}

One may wonder under what scenarios the region proposal methods but ours fail to hit the ground truth regions. Figure~\ref{fig:edgebox} gives some insights by showing the Edgebox region candidates on ReferItGame. We find that the region candidates mainly fail to hit stuff ground truth regions (e.g.,  the ``grass'' in Figure~\ref{fig:edgebox} (c)). Tiny objects are also hard to hit (cf.\ Figure~\ref{fig:edgebox} (e) and (f)). Finally, when a query refers to more than one objects, it might fail region proposal methods which are mostly designed to place a tight bounding box around only one object (cf.\ Figure~\ref{fig:edgebox} (a) and (b)). 

\begin{table}[t]
\centering
\caption{Hit rates of region proposal methods. 
}
\vspace{-0.0in}
\begin{tabular}{ | l | c | c | c | c | }
    \hline
    \multirow{2}{*}{Hit rate, N=200} &
    \multicolumn{2}{|c|}{\small Flickr30K Entities} &
    \multicolumn{2}{|c|}{ReferitGame}\\
    \cline{2-5}
      & val set & test set & val set & test set\\
    \hline
     MRCN Detect.~\cite{he2017mask} & 48.76 & 49.28 & 27.63 & 28.12 \\
     MRCN RP~\cite{he2017mask} & 76.40 & 76.60 & 44.80 & 46.50 \\
     Edgebox~\cite{zitnick2014edge}   & 82.91 & 83.69 & 68.62 & 68.26 \\
     Selec. Search~\cite{uijlings2013selective}  & 84.85 & 85.68 & 81.67 & 80.34 \\
     Ours & {\bf 95.32} & {\bf 95.48} & {\bf 92.40} & {\bf 91.32}  \\
    \hline
\end{tabular}
\vspace{-10pt}
\label{table:oracle}
\end{table}
\vspace{-2pt}
\subsection{Qualitative results analyses}
\vspace{-2pt}
\begin{figure*}[t]
\begin{center}
   \centerline{\includegraphics[width=15.5cm]{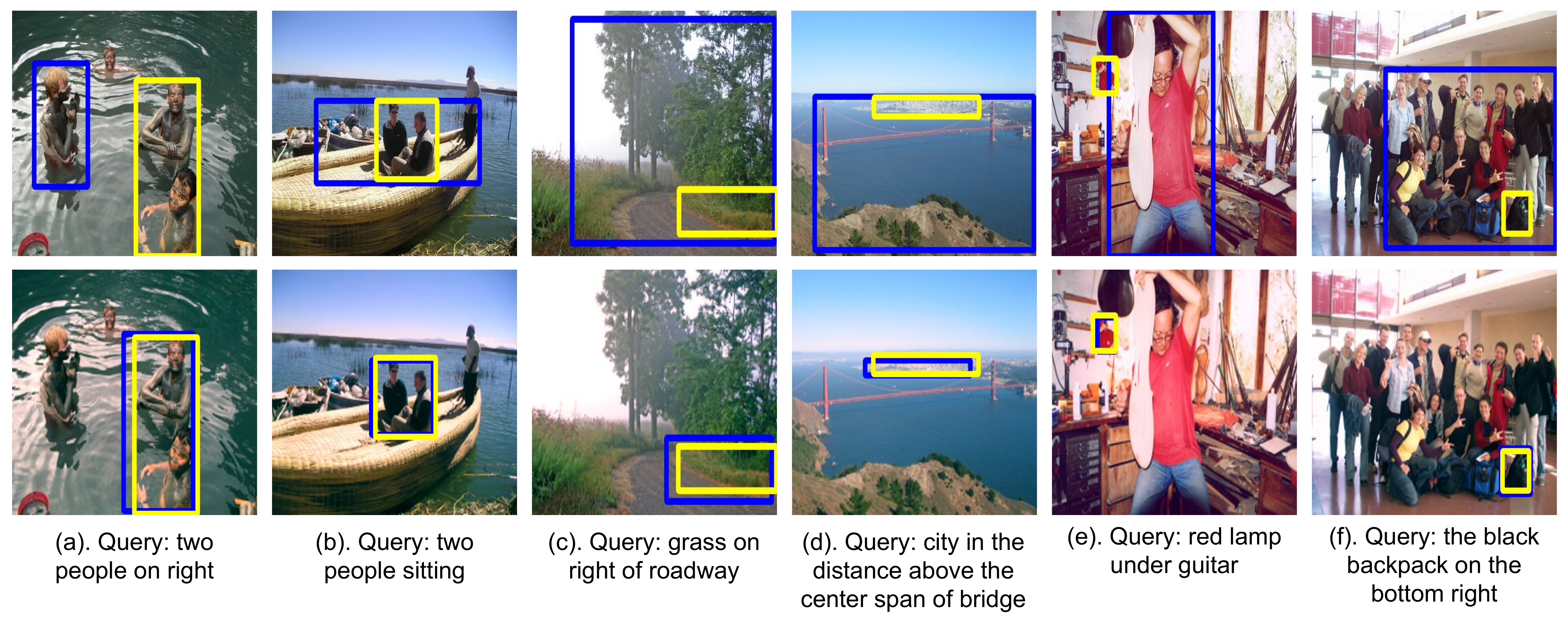}}
\end{center}
\vspace{-0.35in}
	\caption{Mistakes made by the two-stage similarity network (top row) that can be corrected by our one-stage approach (bottom row). Blue boxes are predicted regions and yellow boxes are the ground truth. There are three types of common failures of the two-stage method: queries referring to multiple objects (a,b), queries referring to stuff regions (c, d), and challenging regions (e, f).}
\label{fig:visucmp}
\vspace{-0.05in}
\end{figure*}
\begin{figure*}[t]
\begin{center}
   \centerline{\includegraphics[width=15.5cm]{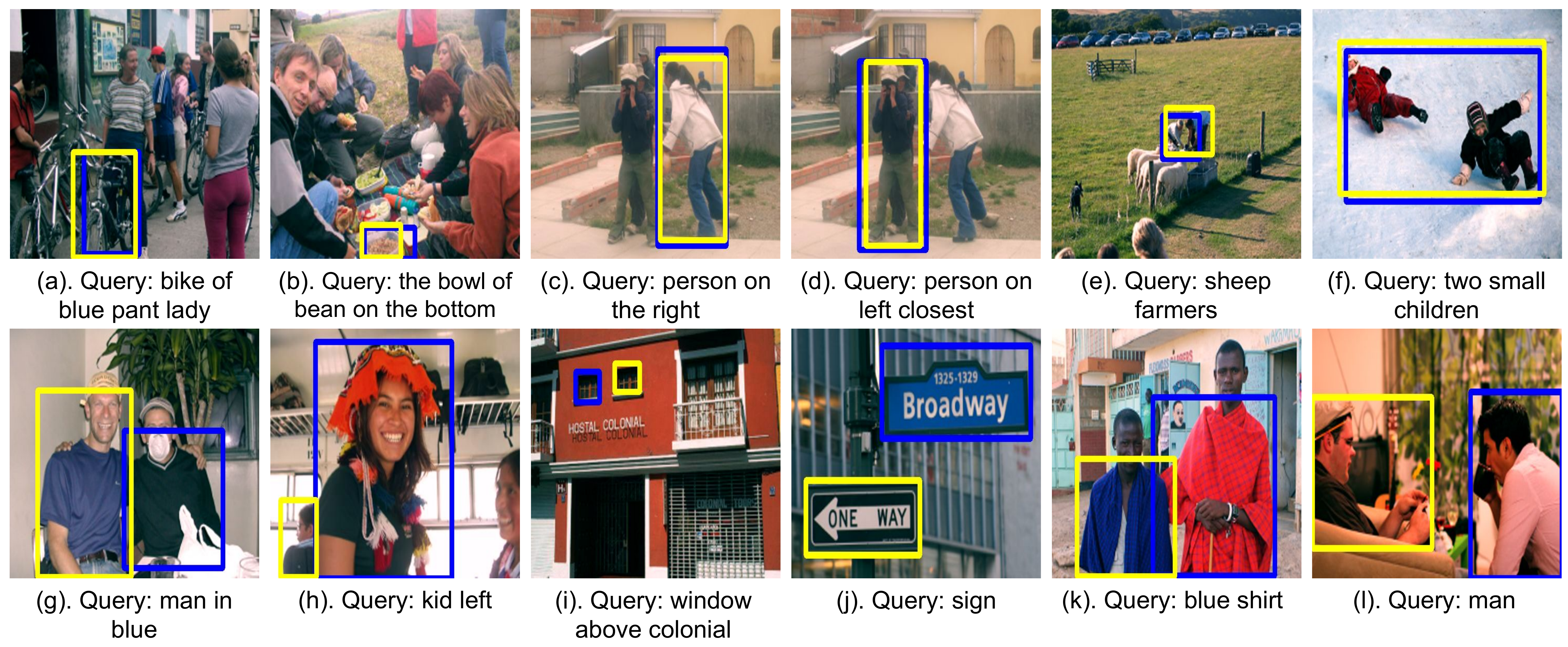}}
\end{center}
\vspace{-0.3in}
	\caption{Success cases on challenging instances (top row) and common failures (bottom row) of our one-stage method. Blue / yellow boxes are predicted regions / ground truths. The four columns on the left are from ReferitGame and the others are from Flickr30K Entities. }
\label{fig:visulist}
\vspace{-0.18in}
\end{figure*}
In this section, we analyze the success and failure cases of the two-stage similarity network as well as our model to show the advantages and limitations of the proposed one-stage method. Figure~\ref{fig:visucmp} shows the mistakes made by the similarity network that can be corrected by our method. The blue boxes are predictions, and the yellow boxes represent the ground truths. We group some common mistakes into the following scenarios.
\vspace{-6pt}
\begin{itemize}
    \setlength\itemsep{-3pt}
    \item {\bf Queries referring to multiple objects.}  A language query in the visual grounding problem can refer to more than one objects, but the region proposals by design aim to each cover only one object. Check the queries ``two people on right'' and ``two people sitting'' in Figures~\ref{fig:visucmp}~(a) and (b), respectively, for examples. it is (almost) impossible to overcome this type of mismatches by the existing propose-and-rank two-stage methods. In contrast, our approach is not restricted to one object per box at all and, instead, can flexibly adapt the box regression function according to the queries.
    \item {\bf Queries referring to stuff as opposed to things.} The second kind of common errors made by the two-stage methods is on the queries referring to stuff as opposed to things,  such like the ``grass'' and ``city'' shown in Figures~\ref{fig:visucmp}~(c) and (d), respectively. This kind of errors is again due to that the region proposals mostly focus on thing classes --- the ``objectness'' is often an important cue for proposing the regions. In sharp contrast, stuff regions generally have low ``objectness'' scores. Hence, we argue that the two-stage methods are incapable of handling such stuff regions given the status quo of region proposal techniques. Our one-stage method can instead learn to handle the stuff regions from the training set of the visual grounding datasets.
    \item {\bf Challenging regions.} In the third kind of common errors, the two-stage methods fail to deal with challenging test cases, such as the small regions referred to by the queries in Figures~\ref{fig:visucmp}~(e) and (f). There are mainly three reasons that fail the two-stage methods. First, the region candidates of the first stage may not provide a good coverage especially over small objects. Second, the visual features of small regions are not discriminative enough for the second stage to learn how to rank. Third, the image depicts complicated scenes or many duplicated objects. The last point could equally harm our approach as well as the two-stage ones.
    \vspace{-6pt}
\end{itemize}

\noindent\textbf{Failure cases of our approach.}
Figure~\ref{fig:visulist} shows extra success and failure cases of our approach. The first row shows the typical success cases. The  ``bike of blue pant lady'' in Figure~\ref{fig:visulist} (a) queries an example image with multiple objects of the same class. Figure~\ref{fig:visulist} (b) provides an example of correct predictions on tiny objects.  (c) and (d) showcase our approach is able to interpret location information in the queries. The query in  (e) contains a distracting noun, ``sheep''. Our model in (f) successfully predicts a region containing two objects.

Figures~\ref{fig:visulist} (g)--(l) are some failure cases of our model. We find our model is insensitive to attributes, such as the ``blue'' in (g) and (k). It fails on (h) and (i) simply because those are very difficult test cases (e.g., one has to recognize the word ``colonial'' in the image in order to make the right prediction). Finally, (j) and (l) give two ambiguous queries for which our model happens to predict different boxes from those annotated by users. 

\section{Conclusion}
We have proposed a simple and yet effective one-stage method for visual grounding. We merge language queries and spatial features into the YOLOv3 object detector and build an end-to-end trainable visual grounding model. It is about 10 times faster than state-of-the-art two-stage methods and achieves superior grounding accuracy. Besides, our analyses reveal that existing region proposal methods are generally not good enough, capping the performance of the two-stage methods and indicating the need of a paradigm shift to the one-stage framework. In future work, we plan to investigate the extensibility of the proposed one-stage framework for modeling other cues in the visual grounding problem.

{\small
\vspace{-3pt}
\subsection*{Acknowledgment}
\vspace{-3pt}
This work is partially supported by NSF awards \#1704337, \#1722847, and \#1813709.
}
\clearpage
{\small
\bibliographystyle{ieee_fullname}
\bibliography{ground}
}
\clearpage
\begin{appendices}
\section{Results on RefCOCO}
The extended experiments on RefCOCO~\cite{yu2016modeling} are reported in Table~\ref{table:refcoco}. RefCOCO contains 19,994 images and 50,000 referred objects originally from MSCOCO~\cite{lin2014microsoft}, with 142,210 collected referring expressions. The referred objects are selected from MSCOCO annotations, and are in one of the 80 object classes defined by MSCOCO with no free-form expressions. We follow the split~\cite{yu2016modeling} of train/validation/testA/testB, which has 120,624, 10,834,
5,657 and 5,095 expressions, respectively. ``testA'' contains images with multiple people and ``testB'' contains images with instances of all other objects.

Organizing the results in the same way as Tables~\ref{table:flickr} and \ref{table:clef}, the state-of-the-art results~\cite{yu2017joint,zhang2018grounding,yu2018mattnet} are reported in the top portion of Table~\ref{table:refcoco}. The middle contains the variants of the similarity network~\cite{wang2019learning}, and the bottom shows our results. Similarly, we list the region proposals used by two-stage methods in the ``Region Proposals'' column. For studies~\cite{yu2017joint,zhang2018grounding,yu2018mattnet} that use different proposals during training and inference, we show the ``Region Proposals'' in a format of ``A/B'' where A stands for the proposals used during training and B during inference. COCO-trained ResNet-101 is used as ``visual features'' and LSTM is used for ``language embedding'' in all reported results unless stated otherwise in method names. 

Ours-LSTM outperforms the state-of-the-art methods except MAttNet~\cite{yu2018mattnet}, which uses extra supervision such as attributes and class labels of region proposals. Compared to the improvements from two-stage methods~\cite{wang2019learning} on ReferitGame (25.0\%) and Flickr30K Entities (7.8\%), the improvement on RefCOCO is rather small (1.4\%). We prove empirically that the major reason is the good proposal quality on RefCOCO. First, the hit rate analyses in Table~\ref{table:oracle-coco} show that the proposals generated by the COCO-trained proposal networks are sufficiently good on RefCOCO. The used proposals cover almost all the referred objects on RefCOCO (92.4\% for detection and 98.5\% for region proposals). This is because the images and objects in RefCOCO are a subset of COCO. Second, among the two-stage methods in the middle of Table~\ref{table:oracle-coco}, the similarity network with COCO-trained Faster R-CNN~\cite{ren2015faster} generated proposals significantly outperforms (by 14.1\%) the one using Edgebox~\cite{zitnick2014edge}. This shows the strong correlation between the good proposal quality and the good performance of two-stage methods on RefCOCO.

Since RefCOCO is a subset of COCO and has shared images and objects, the proposal quality with COCO-trained proposal networks nearly perfect as shown in Table~\ref{table:oracle-coco}. With such ideal proposals on RefCOCO, two-stage methods can greatly narrow the performance gap between one- and two-stage methods. However, this special case only holds on the subset of COCO. Both the hit rate and grounding accuracy drop dramatically when such proposal networks are directly used on other datasets~\cite{kazemzadeh2014referitgame,plummer2017flickr30k}. Table~\ref{table:oracle} reports a lower hit rate of region proposal networks (RPNs) generated proposals compared to Edgebox, which is contradictory to the analyses on RefCOCO in Table~\ref{table:oracle-coco}. Similarly, two-stage methods with RPNs generated proposals perform worse than those with Edgebox. On ReferItGame, similarity network with RPNs generated proposals generates an accuracy of 27.1\%, compared to the 34.5\% with Edgebox. This posts a caveat of the RefCOCO datasets that free-form expressions might be necessary. We hope future works will experiment both with and beyond COCO.

Regarding the problem of imperfect region candidates in two-stage methods, a natural idea is end-to-end fine-tuning region proposal networks (RPNs), which does boost the two-stage methods' overall performances. QRC Net~\cite{chen2017query} trains RPNs in an end-to-end manner and achieves the following results on ReferitGame (Sim. Net~\cite{wang2019learning}: 34.54\%, QRC Net~\cite{chen2017query}: 44.07\%, Ours: 59.30\%) and Flickr30K Entities (Sim. Net~\cite{wang2019learning}: 60.89\%, QRC Net~\cite{chen2017query}: 65.14\%, Ours: 68.69\%). Besides, the two-stage methods perform better on RefCOCO (cf.\ Tables~\ref{table:refcoco} and~\ref{table:oracle-coco}) than them on the other two datasets because their RPNs are trained not only by the mentions in the referring expressions but also other COCO objects. Nonetheless, our approach still gives rise to better overall results (as well as the faster inference speed and simpler framework).

\begin{table}[t]\fontsize{8}{9}\selectfont
\centering
\caption{Referring expression comprehension results on RefCOCO~\cite{yu2016modeling}. LSTM and COCO-trained Res101 are the encoders unless stated otherwise in method names.}
\vspace{0.03in}
\begin{tabular}{ l l c c c }
    \hline
    Method & Region Proposals & val & testA & testB\\
    \hline
    SLR~\cite{yu2017joint} & GT/FRCN Detc. & 69.48 & 73.71 & 64.96 \\
    VC-VGG16~\cite{zhang2018grounding} & GT/SSD Detc.~\cite{liu2016ssd} & - & 73.33 & 67.44 \\
    MAttNet Base~\cite{yu2018mattnet} & GT/FRCN Detc. & 72.72 & 76.17 & 68.18 \\
    MAttNet~\cite{yu2018mattnet} & GT/FRCN Detc. & {\bf 76.40} & {\bf 80.43} & 69.28 \\
    \hline
    Similarity Net~\cite{wang2019learning} & Edgebox N=200 & 57.33 & 57.22 & 55.60\\
    Similarity Net~\cite{wang2019learning} & GT/FRCN Detc. & 71.48 & 74.90 & 67.32 \\
    Sim. Net-Darknet~\cite{wang2019learning} & GT/FRCN Detc. & 72.27 & 75.12 & 67.91 \\
    \hline
    Ours-Darknet-Bert & None & 72.05 & 74.81 & 67.59 \\
    Ours-Darknet-LSTM & None & 73.66 & 75.78 & {\bf 71.32} \\
    \hline
\end{tabular}
\vspace{-0.1in}
\label{table:refcoco}
\end{table}
\begin{table}
\centering
\caption{Hit rates of region proposals on RefCOCO.}
\vspace{-0.0in}
\begin{tabular}{ | l | c | c | c | }
    \hline
    Hit rate, N=200 & val & testA & testB\\
    \hline
     FRCN Detc.~\cite{ren2015faster} & 92.42 & 95.83 & 88.87 \\
     FRCN RP~\cite{ren2015faster} & 98.52 & {\bf 99.47} & 97.60\\
     Edgebox~\cite{zitnick2014edge} & 89.01 & 89.62 & 89.28\\
     SS~\cite{uijlings2013selective} & 84.28 & 81.72 & 89.54\\
     Ours & {\bf 98.80} & 99.08 & {\bf 98.64} \\
    \hline
\end{tabular}
\label{table:oracle-coco}
\end{table}

\section{Cross-Sample Relationships}
\begin{table}[t]\fontsize{8}{9}\selectfont
\centering
\caption{Visual grounding results of cross-sample methods.}
\vspace{-0.0in}
\begin{tabular}{ l c c c c c }
    \hline
    \multirow{2}{*}{Method} & ReferIt & Flickr30K &
    \multicolumn{3}{c}{RefCOCO}\\
     & Game & Entities & val & testA & testB \\
    \hline
    SeqGROUND~\cite{dogan2019neural} & - & 61.60 & - & - & - \\
    QRC Net~\cite{chen2017query} & 44.07 & 65.14 & - & - & - \\
    \hline
    Ours                & 59.30 & 68.69 & 73.66 & 75.78 & 71.32 \\
    Ours-Cross sample & 60.37 & 69.15 & 74.52 & 76.51 & 71.88 \\
    \hline
\end{tabular}
\vspace{-0.1in}
\label{table:reg}
\end{table}
Inspired by previous studies~\cite{chen2017query,dogan2019neural} that successfully exploit all phrases and queries on the same image for visual grounding, we extend our vanilla framework to utilize cross-sample relationships. Given an anchor sample with image $I_i$ and query $Q_i$ describing an object of interest, we define samples in the positive bag as all other pairs with different queries describing the same object in the same image. For example in Figure~\ref{fig:arch}, ``two people sitting'' and ``two people in the middle of the boat'' refer to the same region. The negative bag consists of the intra-image negative samples with the queries describing different regions in the same image, and the inter-image negative samples with completely different images. 
Given an anchor sample, we assume that the fused visual-textual feature should be more similar to the ones in the positive bag compared to the negative ones. The proposed feature regularization enforces such relationship with a triplet loss:
\begin{align*}
L_{reg} &= \sum_{i} \left[ \lVert f(I_i, Q_i) - f(I_{P_i}, Q_{P_i}) \rVert_2^2 \right . \\
& - \left . \lVert f(I_i, Q_i)-f(I_{N_i}, Q_{N_i})\rVert_2^2 + m \right ]_+
\end{align*}
where ${(I_{P_i}, Q_{P_i})}$ and ${(I_{N_i}, Q_{N_i})}$ are the sampled positive and negative image-query pairs. The feature $f$ can be any fused visual textual feature. In this study, we define $f$ as the average pooling results of the feature in the last but one layer.

In experiments, we set margin $m=1$ and regularization term weight $w_{reg}=1$. Table~\ref{table:reg} reports the performance with feature regularization. We observe an improvement in performance on all three datasets~\cite{kazemzadeh2014referitgame,plummer2017flickr30k,yu2016modeling}. We also experiment with hard triplet generation, but observe no major change in performances.

\end{appendices}
\end{document}